%% file: main.tex
\documentclass[letterpaper, 10 pt, conference]{ieeeconf}  
\IEEEoverridecommandlockouts                              
\pdfoutput=1
\usepackage{tabularray}
\usepackage{graphicx}
\usepackage{epstopdf}
\usepackage[utf8]{inputenc}
\usepackage{pgfplots}
\usepackage{amsmath,amssymb}
\usepackage{subfig}
\usepackage{xfrac}%for slant fraction
\usepackage{algorithm}
\usepackage[noend]{algpseudocode}
\usepackage{bm}
\usepackage{wrapfig}
\usepackage{float}
\usepackage{orcidlink}
\floatstyle{plaintop}
\restylefloat{table}

\usepackage{cite}

\pgfplotsset{compat=1.17}

\newtheorem{definition}{Definition}

\usepackage{xcolor}

\title{System Identification and Control of Front-Steered Ackermann Vehicles through Differentiable Physics}

\author{Burak M. Gonultas$^{1}$\orcidlink{0000-0002-7966-7929}, Pratik Mukherjee$^{1}$\orcidlink{0000-0003-2970-8515}, O. Goktug Poyrazoglu$^{1}$\orcidlink{0009-0002-3778-100X} and Volkan Isler$^{1}$\orcidlink{0000-0002-0868-5441}% <-this % 
\thanks{$^{1}$ the authors are with the Department of Computer Science and Engineering, University of Minnesota, Minneapolis, MN, 55455, USA
        {\tt\small \{gonul004, mukhe027, poyra002, isler\}@umn.edu}}%
}
\usepackage{fancyhdr}
\fancypagestyle{firststyle}
{
\fancyhf{} 
\fancyfoot[CO]{\copyright 2023 IEEE. Personal use is permitted, but republication/redistribution requires IEEE permission.
See \href{https://www.ieee.org/publications/rights/index.html}{https://www.ieee.org/publications/rights/index.html}  for more information.}
}
\begin{document}
\maketitle
\thispagestyle{firststyle}
\input{0-abstract}
\input{1-introduction.tex}

\input{2-technical_background.tex}
\input{3-problem_formulation}
\input{4-proposed_method}

\input{5-simulation_results}

\input{6-generalization_experiments.tex}

\input{7-Conclusion_and_future_directions}

\bibliographystyle{IEEEtran}
\bibliography{reference}
\end{document}

%% file: 0-abstract.tex
\begin{abstract}

In this paper, we address the problem of system identification and control of a front-steered vehicle which abides by the Ackermann geometry constraints.
This problem arises naturally for on-road and off-road vehicles that require reliable system identification and basic feedback controllers for various applications such as lane keeping and way-point navigation. Traditional system identification requires expensive equipment and is time consuming. In this work we explore the use of differentiable physics for system identification and controller design and make the following contributions: i)~We develop a differentiable physics simulator (DPS) to provide a method for the system identification of front-steered class of vehicles whose system parameters are learned using a gradient-based method; ii) We provide results for our gradient-based method that exhibit better sample efficiency in comparison to other gradient-free methods; iii) We validate the learned system parameters by implementing a feedback controller to demonstrate stable lane keeping performance on a \textit{real} front-steered vehicle, the F1TENTH;
 iv) Further, we provide results exhibiting comparable lane keeping behavior for system parameters learned using our gradient-based method with lane keeping behavior of the \textit{actual} system parameters of the F1TENTH.

\end{abstract}

%% file: 1-introduction.tex
\section{Introduction}
Performing experiments with real robots is a difficult, time-consuming issue and often a costly task.
To address these issues, there has been growing interest in the robotics community to reduce the gap between simulation and real-world experiments. The intention is to provide roboticists with a virtual environment to develop algorithms and test them on robots in a risk-free manner. However, the trade-off is that validating the performance of algorithms in simulation often does not directly translate to the real-world robots. This primarily occurs because readily available simulation environments are low-fidelity. 
%because much of existing simulation environments are not able to emulate a real-world scenario for roboticists to validate their work reliably. 
% Further, when it comes to testing off-the-shelf controllers on real robots, what works in high fidelity simulation environment, does not always work on the real robotic system. The primary reason is again the sim-to-real gap \cite{neunert2017off}. 
Current literature have explored two different approaches to tackle the sim-to-real gap issue: i) Developing accurate models of robots and improving the fidelity of simulation environments; ii) Making robot controllers more robust and adaptive to uncertainty in the environment. Source code is available on our
project page: \href{https://github.com/gonultasbu/diffsteered}{\color{magenta}{https://github.com/gonultasbu/diffsteered}}
% \vtxt{I don't understand the point of this paragraph. Is it that, ``even if the simulation was high fidelity, decoupling system identificatino from controller design causes performance issues"? }

Conventionally, improvement on the system model is achieved by traditional \textit{system identification}. Traditional methods for system identification \cite{sa2018dynamic} include the \textit{frequency} and \textit{impulse} response methods, where the system parameter is obtained offline using predefined input signals. This typically is a time consuming endeavour and requires the presence of real system which can be expensive and is not always available. 
% For instance, to conduct system identification on a fighter jet using the traditional methods, one needs to own a real fighter jet or a very expensive simulator. 
Researchers have also developed online methods for conducting system identification using the recursive least squares method for identifying linear systems \cite{kaess2008isam} and local linear regressors for identifying affine time-varying models \cite{rosolia2019learning}. Recently, learning-based approaches have also been applied to improve model fidelity. For instance, \cite{granados2022model} performs model identification of mobile \textit{skid-steered} robots  using a differentiable physics simulation environment. Our work in this paper extends the work in \cite{granados2022model} to front-steered Ackermann type vehicles.

\begin{figure}[!htb]
\centering
% \subfloat{\includegraphics[width = 0.8\linewidth]
\subfloat{\includegraphics[width=0.48\textwidth]
{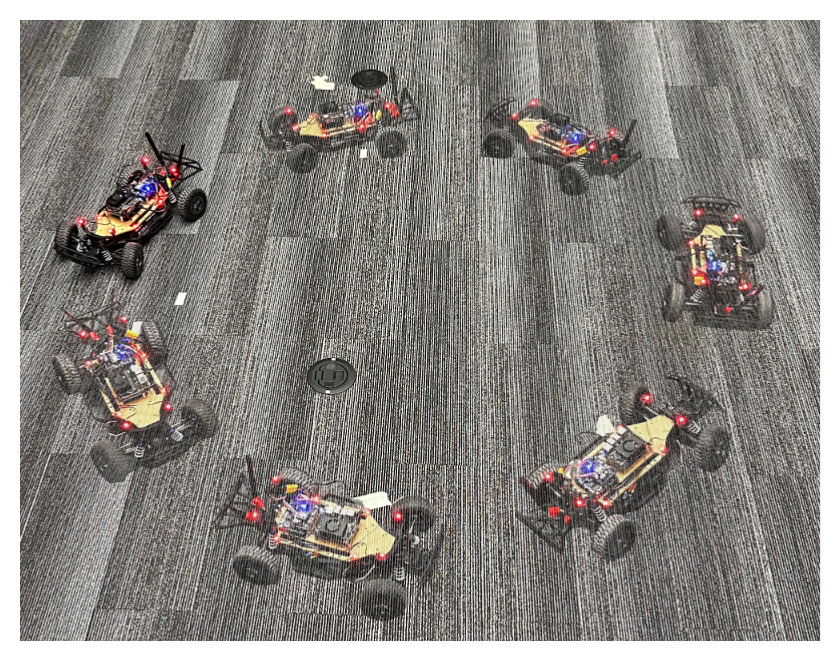}}
\caption{A feedback controller exhibiting a lane keeping behavior for the F1TENTH robot to track a circular trajectory. }
% For every \textit{one epoch}, the pole placement method is used to determine the stable controller gain $K$, using the identified system parameter $(A,B)$, which is then run for $\mathcal{T}$ timesteps in DPS environment with the control law $\delta_t$ for each time step $t$ to get the propagated system state $s_{t+1}$ and compute the accumulated loss $\mathcal{L}$ based on the lateral distance error state $x_t$ from the lane centerline.} 
\vspace{-10pt}
\label{fig:dps_framework}
\end{figure}

On the other hand, when it comes to using controllers to reduce the sim-to-real gap, researchers have worked on developing \textit{robust} and \textit{adaptive} controllers.
Hovakimyan and Cao~\cite{hovakimyan201011} provide a detailed study consisting of the traditional adaptive controllers, as well as modern adaptive controllers that are not only \textit{stable} but also \textit{robust}. However, designing a robust controller that is adaptive by nature requires the design of Lyapunov functions that do not necessarily cater to general systems. Moreover, deriving adaptive controllers and implementing them on real robots is not trivial.

In this work, we focus on the class of mobile robots with front wheel steering. This is an important class of robots which consists of everyday on-road and off-road vehicles used in various applications, i.e., farming, mining, etc. Moreover, with an increase in interest for autonomous vehicles, we realized that there are not many works 
\cite{yu2017simultaneous,carpin2007usarsim,boer2002physical} that tackle the problem of system identification and control of front-steered mobile robots, and especially using \textit{differentiable physics simulators} (DPS) \cite{hu2019difftaichi, freeman2021brax, rohrbein2021differentiable, heiden2019interactive, de2018end, degrave2019differentiable, geilinger2020add, qiao2020scalable, toussaint2018differentiable, schenck2018spnets, liang2019differentiable, hu2019chainqueen}. From a controls perspective, there is extensive literature that is provided by the controls community starting with \cite{rajamani2011vehicle}. In \cite{rajamani2011vehicle}, Rajamani details \textit{basic} off-the-shelf controllers such as feedback controllers for front-steered vehicles which abide by the Ackermann geometry constraints \cite{ackermann1997robust} for the application of lane keeping. Similarly, from a system identification perspective, for mobile robots, recent works such as \cite{granados2022model,chen2022real} have addressed solving \textit{simultaneous} system identification and control, but specifically for \textit{skid-steered} vehicles and manipulator type robotic systems, respectively. Realizing this void for system identification and control of front-steered vehicles in the robotics community, which serve a vast array of applications in real-world, we are motivated to provide a robust framework, as shown in Fig. \ref{fig:dps_framework}, using DPS for system  identification and control of front-steered vehicles. The framework in Fig. \ref{fig:dps_framework} depicts the system parameter learning using gradient based optimization methods (grey region) offline. Once the parameters are learned in the form of system dynamics equation matrices $(A,B)$, we validate the learned parameters using a \textit{feedback} controller for the application of lane keeping of front-steered Ackermann vehicles (orange region). In this case, we conduct the \textit{system identification} of a readily available front-steered vehicle, the F1TENTH, and validate a \textit{traditional} feedback controller for lane keeping application. The F1TENTH \cite{o2020f1tenth} is a widely used mobile robot in the robotics community for demonstrating autonomous vehicle applications.

\begin{figure}[!htb]
\centering
% \subfloat{\includegraphics[width = 0.8\linewidth]
\subfloat{\includegraphics[width=0.42\textwidth]
{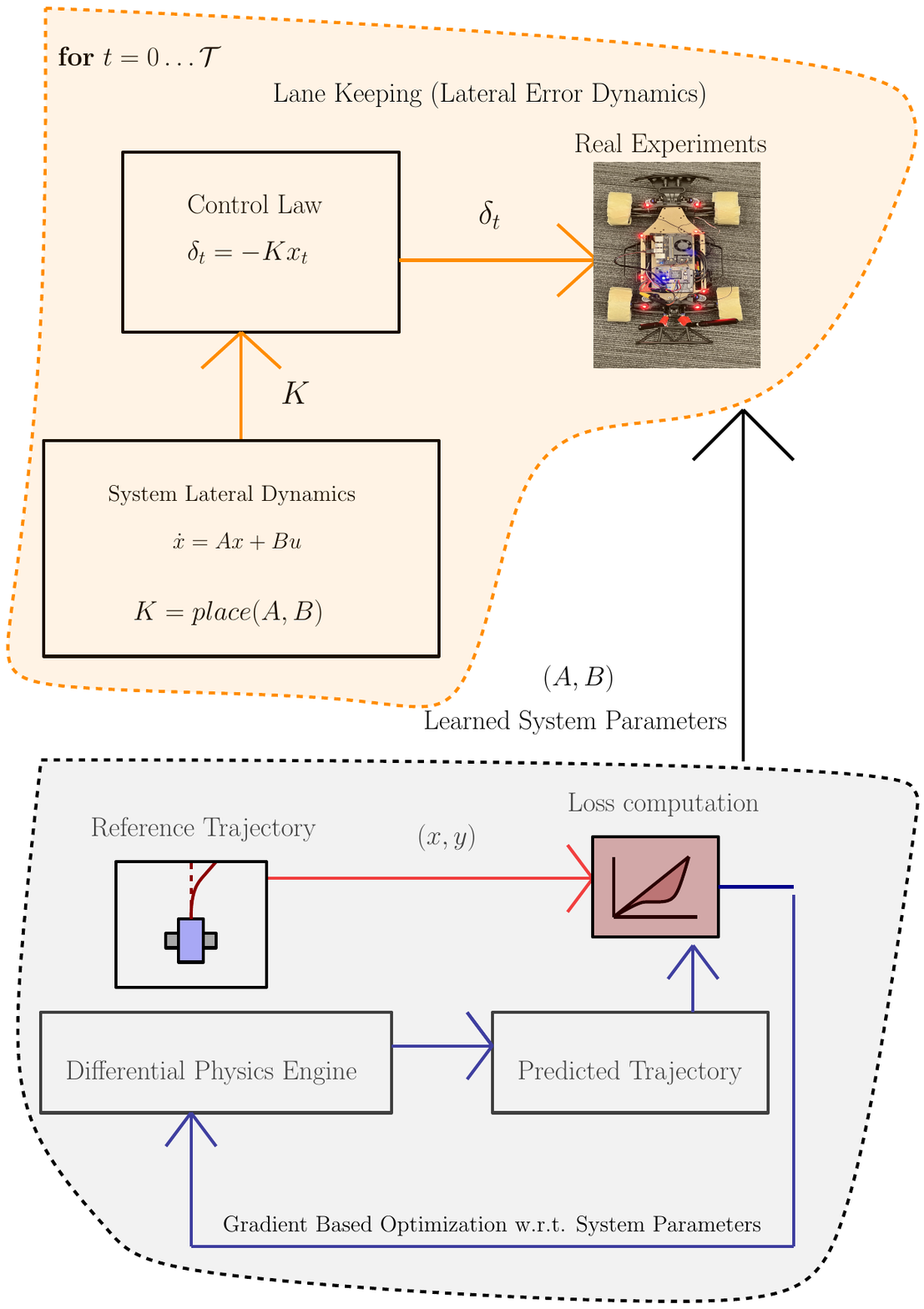}}
\caption{Overview of our approach depicts offline system identification (grey region) for general systems and feedback control (orange region) for lane keeping application of front-steered F1TENTH vehicle. }
% For every \textit{one epoch}, the pole placement method is used to determine the stable controller gain $K$, using the identified system parameter $(A,B)$, which is then run for $\mathcal{T}$ timesteps in DPS environment with the control law $\delta_t$ for each time step $t$ to get the propagated system state $s_{t+1}$ and compute the accumulated loss $\mathcal{L}$ based on the lateral distance error state $x_t$ from the lane centerline.} 
\vspace{-10pt}
\label{fig:dps_framework}
\end{figure}
In this regard, our contributions are fourfold:
% \vtxt{say something about why this is non-trivial} 
i) We use differentiable physics simulator (DPS) to provide a method for the system identification of the front-steered F1TENTH vehicle 
% \vtxt{what does it mean to ``address" as a contribution? Do we provide a method, algorithm, equations etc?}
 whose system parameters are learned using a gradient-based method; ii) We provide results for our gradient-based method that exhibit better sample efficiency in comparison to other gradient-free methods; iii) We validate the learned system parameters by implementing a feedback controller to demonstrate stable lane keeping performance on a \textit{real} front-steered vehicle, the F1TENTH;
  iv) Further, we provide results exhibiting comparable lane keeping behavior for system parameters learned using our gradient-based method with lane keeping behavior of the \textit{actual} system parameters of the F1TENTH.

% We address the system identification and control of a front-steered vehicle, which satisfies the Ackermann geometry constraints, as a coupled problem; iii) We demonstrate lane keeping using a general feedback controller for a front-steered vehicle whose system parameters as well as the controller gains are learned using a gradient-based method only enabled by a DPS; iv) We provide extensive simulation results validating our methods by comparing controller performance for a front-steered vehicle, whose parameters are identified using DPS, on other off-the-shelf non-DPS environments. 
% \vtxt{see my comments in the abstract}
% \ptxt{discuss why simulations suffice+ very with controllers}

The rest of the paper is organized as follows. Technical background
material is summarized in
Section \ref{Technical Background}. The formal problem formulation is given in Section \ref{Problem Formulation}. In Section \ref{Proposed Method}, we provide extensive details on our method for system identification. Finally, we provide simulation results in Section \ref{Simulation Results} and experimental results with F1TENTH in Section\ref{Generalization Experiment Results}
and conclude the paper in Section \ref{Conclusions and Future Directions}.

% \ptxt{need high level block diagram for complete framework explaining system ID + control}

%% file: 2-technical_background.tex
\section{Technical Background}\label{Technical Background}
In this section, we briefly review the theory behind the front-steered vehicle lateral \textit{error} dynamics derived in \cite{rajamani2011vehicle} and discuss the framework of DPS used for our simulations.
\subsection{Vehicle lateral Dynamics}
\begin{figure}[!htb]
\centering
\subfloat{\includegraphics[ width=8cm,height=8cm]{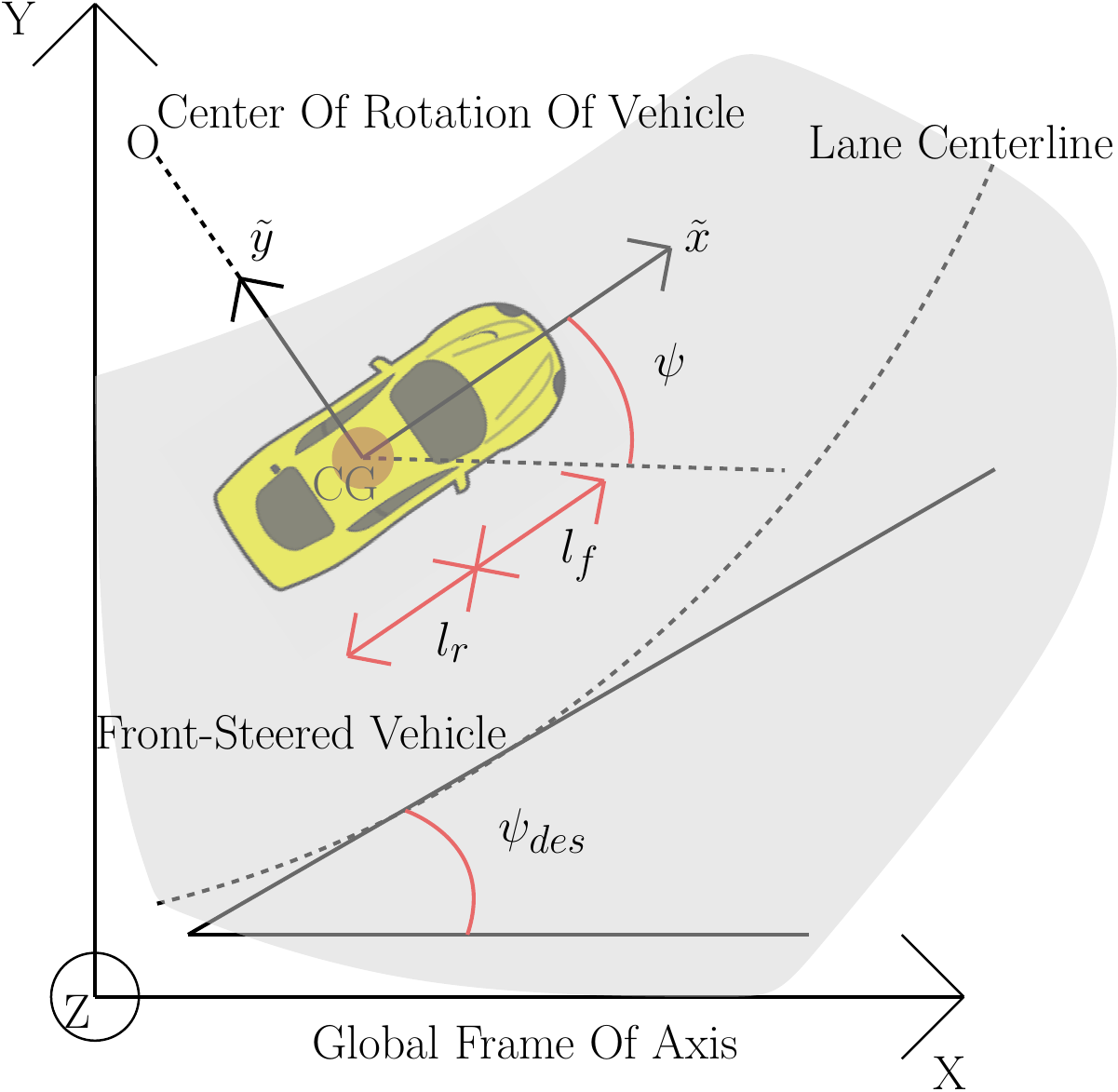}}
\caption{A lateral lane keeping system for a front-steered vehicle.}
\label{fig:lane_keeping}
\end{figure}
 Fig. \ref{fig:lane_keeping} shows a simplified linear two-degrees of freedom (2-DOF) bicycle  model of the vehicle lateral dynamics derived in \cite{rajamani2011vehicle}.
 % The original model for the vehicle lateral dynamics has sixteen state variables: twelve for the six DOF motions (three translation and three rotational) and four for the tires. A simplified linear two-degree of freedom bicycle  model of the vehicle is obtained from the complex model and as shown in Fig.\ref{fig:lane_keeping} will be used in this work. 
 2-DOF are $\psi$, the vehicle yaw angle, and $y$, the lateral position with respect to the center of the rotation of the vehicle $O$.  The yaw angle is considered as the angle between horizontal \textit{vehicle body frame} axis of the vehicle, $(\Tilde{x})$, and the global horizontal axis, $(X)$. The \textit{constant} longitudinal velocity of the vehicle at its center of gravity ($\boldsymbol{CG}$) is denoted by $V_x$ and the mass of the vehicle is denoted by $m$. The distances of the front and rear tires from the $\boldsymbol{CG}$ are shown by $l_f$, $l_r$, respectively, and the front and rear tire cornering stiffness are denoted by $C_{af}$ and $C_{ar}$, respectively. The steering angle is denoted by $\delta$ which also serves as the control signal when a controller is implemented and the yaw moment of inertia of the vehicle is denoted by $I_z$. Considering the lateral position, yaw angle, and their derivatives as the state variables, and using the Newton's second law for the motion along the \textit{vehicle body frame} $\Tilde{y}-axis$, the state space model of lateral vehicle dynamics is derived in \cite{rajamani2011vehicle}. 
 % For the sake of brevity we do not discuss the state space model of the lateral vehicle dynamics here.
 
%   \begin{align} \label{eq:lat_dyn_st_sp}
%  \underbrace{\frac{d}{dt}\begin{bmatrix}
% y  \\
% \dot{y} \\
% \psi\\
% \dot{\psi} 
% \end{bmatrix}}{\dot{x}} &= 
% \begin{bmatrix}
% 0 & 1 & 0 & 0\\
% 0 & -\frac{2 C_{af}+2 C_{ar}}{m V_x} & 0 & -V_x - \frac{2 C_{af}l_f-2 C_{ar}}{m V_x} l_r\\
% 0 & 0 & 0 & 1\\
% 0 & -\frac{2 C_{af}l_f-2 C_{ar}l_r}{I_z V_x} & 0 & -\frac{2 C_{af}l_f^2+2 C_{ar}l_r^2}{I_z V_x} 
% \end{bmatrix} \nonumber\\
% &\times 
% \begin{bmatrix}
% y  \\
% \dot{y} \\
% \psi\\
% \dot{\psi} 
% \end{bmatrix} + \begin{bmatrix}
% 0  \\
% \frac{2 C_{af}}{m} \\
% 0\\
% \frac{2C_{af}l_f}{I_z}
% \end{bmatrix}\delta
% \end{align}
% Since we are primarily concerned with the lane keeping application and for the sake of brevity, we only provide the state-space representation of lateral-dynamics to a model where the state variables are the position and orientation errors with respect to the center of the lane. 
The error dynamics is written with two error variables : $e_1$, which is the distance between the $\boldsymbol{CG}$ of the vehicle from the center line of the lane; $e_2$, which is the orientation error of the vehicle with respect to the desired yaw angle $\psi_{des}$.
Assuming the radius of the road is $R$, the rate of change of the desired orientation of the vehicle can be defined as $\dot{\psi}_{des}= \frac{V_x}{R}$.
% \begin{align}\label{eq:des_orn}
%     \dot{\psi}_{des}= \frac{V_x}{R}
% \end{align}
The tracking or lane keeping objective of the lateral control problem is expressed as a problem  of stabilizing the following \textit{error} dynamics at the origin.
% \resizebox{.9\hsize}{!}{$A+B+C+D+E+F+G+H+I+J+K+L+M+N+O+P+Q+R+S+T+U+V+W+X+Y+Z$}
 \begin{align} \label{eq:er_dyn_st_sp}
&  \underbrace{\frac{d}{dt}\begin{bmatrix}
e_1  \\
\dot{e_1} \\
e_2\\
\dot{e_2} 
\end{bmatrix}}_{\dot{x}} =
\resizebox{0.75\hsize}{!}{  $\underbrace{\begin{bmatrix}
0 & 1 & 0 & 0\\
0 & -\frac{2 C_{af}+2 C_{ar}}{m V_x} & \frac{2 C_{af}+2 C_{ar}}{m } &  - \frac{2 C_{af}l_f-2 C_{ar}}{m V_x} l_r\\
0 & 0 & 0 & 1\\
0 & -\frac{2 C_{af}l_f-2 C_{ar}l_r}{I_z V_x} & \frac{2 C_{af}l_f-2 C_{ar}l_r}{I_z } & -\frac{2 C_{af}l_f^2+2 C_{ar}l_r^2}{I_z V_x} 
\end{bmatrix}}_{A}$} && \nonumber\\
& \times 
\underbrace{\begin{bmatrix}
e_1  \\
\dot{e_1} \\
e_2\\
\dot{e_2} 
\end{bmatrix}}_{x} + \underbrace{\begin{bmatrix}
0  \\
\frac{2 C_{af}}{m} \\
0\\
\frac{2C_{af}l_f}{I_z}
\end{bmatrix}}_{B_1}\underbrace{\delta}_{u} + \underbrace{\begin{bmatrix}
0  \\
-V_x - \frac{2 C_{af}l_f-2 C_{ar}}{m V_x} \\
0\\
-\frac{2 C_{af}l_f^2+2 C_{ar}l_r^2}{I_z V_x}
\end{bmatrix}}_{B_2} \dot{\psi}_{des}
\end{align}
 
Therefore, the above state-space form in \eqref{eq:er_dyn_st_sp} can be represented in the \textit{general} state space form as 

\begin{align} \label{eq:er_gnrl_st_sp}
\dot{x}(t) = Ax(t) + B_1(u(t)) + B_2 \dot{\Psi}_{des}
\end{align}
where $x(t)\in \mathbb{R}^n$, $t \geq 0$ is a state vector, $u(t) \in \mathbb{R}^m$, $t \geq 0$ is the control input which in this case is the steering angle $\delta$. The elements of $A \in  \mathbb{R}^{n\times n}$, $B_1, B_2 \in  \mathbb{R}^{n\times m}$ matrices are generally considered known, but in our case we obtain the elements of matrices using system identification and we assume pair $(A,B_1)$ is controllable. We also provide the definition of a closed-loop system.
\begin{definition}\label{def:closed_loop}
    The closed-loop system is defined as the system described by equation \eqref{eq:er_gnrl_st_sp}, where $u(t)=\delta_t, \forall t$, is the steering angle input to the system generated by a feedback controller .
\end{definition}

\subsection{Differentiable Physics Simulator (DPS)}
Differentiable physics simulators provide analytical gradients for physical systems, which may be used for learning-based solutions such as solving inverse problems, system identification and controller design. Enabled by the recent developments in automatic differentiation literature \cite{paszke2019pytorch, abadi2016tensorflow, al2016theano, hu2019taichi, bell2012cppad, bradbury2018jax}, a number of differentiable simulators have been proposed \cite{hu2019difftaichi, freeman2021brax, rohrbein2021differentiable, heiden2019interactive, de2018end, degrave2019differentiable, geilinger2020add, qiao2020scalable, toussaint2018differentiable, schenck2018spnets, liang2019differentiable, hu2019chainqueen}. While these simulators all aim to provide analytical gradients for learning-based methods, the variance in terms of supported features is high. In this paper, PyTorch \cite{paszke2019pytorch} library is used to develop the differentiable physics engine, which is differentiable with respect to the dynamic model parameters and control inputs of front-steered Ackermann vehicles.

The dynamic model, which is represented by a state space model as shown in \eqref{eq:adp_vdot}, consists of the following states $x_1=s_x, x_2=s_y,x_3=~\delta, x_4=v, x_5= \psi, x_6=\dot{\psi}, x_7=\beta$.
\begin{equation}\label{eq:adp_vdot}
\begin{aligned}
&\dot{x_1} = x_4 cos (x_5 +x_7)\\
&\dot{x_2} = x_4 sin (x_5+x_7)\\
&\dot{x_3} = f_{steer}(x_3,u_1)\\
&\dot{x_4} = f_{acc}(x_4,u_2)\\
&\dot{x_5} = x_6\\
&\dot{x_6} = \frac{\mu m}{I_z (l_r+l_f)}\big (l_f C_{s,f}(gl_r-u_2h_{cg})x_3 + (l_rC_{S,r} \dots\\
&(gl_f+u_2h_{cg})-l_fC_{S,f}(gl_r-u_2h_{cg}))x_7 \dots \\
&-(l^2_f C_{S,f}(gl_r-u_2h_{cg})x_3) + l^2_r C_{S,r}(gl_f+u_2h_{cg}))\frac{x_6}{x_4}   )\\
&\dot{x_7} = \frac{\mu }{x_4 (l_r+l_f)}\big ( C_{s,f}(gl_r-u_2h_{cg})x_3 - (C_{S,r} \dots\\
&(gl_f+u_2h_{cg})+C_{S,f}(gl_r-u_2h_{cg}))x_7 \dots \\
&+( C_{S,r}(gl_f-u_2h_{cg})l_r) -  C_{S,f}(gl_r-u_2h_{cg})l_f)\frac{x_6}{x_4}   ) -x_6
\end{aligned}    
\end{equation}
where $s_x$ is the $x$ position in global coordinates in meters, $s_y$ is the $y$ position in global coordinates in meters, $v$ is the longitudinal velocity in m/s, and $\beta$ is the slip angle at the vehicle center in radians. Defined vehicle parameters are as follows: 
% $m$ is the total vehicle mass in $kg$, $I_{z}$ is the moment of inertia for entire mass about the z-axis in $kg.m^{2}$, $l_{f}$ and $l_r$ are the distance from center of gravity to the front and rear axles, respectively in meters. 
$h_{cg}$ is the center of gravity height of total mass in meters. $\mu$ is the friction coefficient, $C_{S,f}$ and $C_{S,r}$ are tire cornering stiffness coefficients for front and rear wheels in $1/rad$. Inputs are defined as $u_{1}$ representing the steering velocity and $u_{2}$ representing the longitudinal acceleration. The vehicle parameters and definitions are compatible between equations \eqref{eq:er_dyn_st_sp} and \eqref{eq:adp_vdot} except for the angles and related coefficients. Eq. \eqref{eq:er_dyn_st_sp} uses degrees instead of radians, therefore cornering stiffness coefficients from Eq. \eqref{eq:adp_vdot} must be converted to cornering stiffness values which are in degrees. The relation between cornering stiffness and the corresponding coefficient is as follows:
\begin{equation}\label{eq:cornering_stiffness}
\begin{aligned}
&C_{ai} = \mu C_{S,i}F_{z,i}
\end{aligned}    
\end{equation}
Where the subscript $i$ assigns a tire or axle to the front and the rear. Therefore, $F_{z,i}$ becomes the vertical force on the front or rear axle in $N$.

%In this paper, we adopt the Nimble physics simulator \cite{werling2021fast} for analytical gradient computation to develop a \textit{feedback} controller and to identify dynamic system parameters for front-steered vehicles through the control law described in Section II. Nimble physics simulator is the most suitable for our purposes because it is a  general differentiable physics simulator providing gradients through the collision mechanics between the necessary rigid body shapes for the design of a front-steered vehicle. We argue that, while it is possible to develop a differentiable simulator solely focusing on vehicle dynamics using the analytical physics equations for a vehicle model from scratch, such simulators would lack generalization capabilities and be less useful for the broader research community overall.

% a number of differentiable physics engines have been proposed to solve system identification and control problems [16, 20–22, 22, 23, 23–29]. In this work, we adopt the Nimble simulation [35] for taking advantage of the differentiation to develop control laws and identify system parameters which offers fast geometric analytical gradients through collision detection algorithms which support various types of 3D geometry and meshes. Support for various types of 3D geometry and joints is essential in order to be able to have an approximate simulation model for front-steered vehicles.

%% file: 3-problem_formulation.tex
\section{Problem Formulation}\label{Problem Formulation}
At a high-level we want to conduct system identification and learn a dynamic model for front-steered vehicles. Therefore, we formulate the system identification problem as a general optimization problem in terms of identified system dynamic parameters in a DPS environment as the decision variables. The optimization goal is to minimize the gap between the trajectory of the real system and the simulated system. To verify the task based sim2real performance of the identified system parameters, we use a \textit{feedback} controller of the type used in \cite{rajamani2011vehicle} as the basic controller for lane keeping with the error dynamics shown in state-space form in Eq. \eqref{eq:er_dyn_st_sp}.

In the following sub-sections we further elaborate on our \textit{problem definition}
% In the following sub-sections we will define our problem formulation in three parts: i)the general task of lane keeping; ii) the tuning aspect of the applied \textit{feedback} controller; iii)  the learning framework for the dynamic model parameters.
\subsection{Task Representation}
Given a front steered vehicle with center of gravity $\boldsymbol{CG}$, its generalized position at time $t$ in real world is defined as a 2-tuple of Cartesian coordinates $(x^{real}_{t},y^{real}_{t})$. Similarly, its generalized position at time $t$ in simulation is defined as $(x^{sim}_{t},y^{sim}_{t})$. In simulation, the front-steered vehicle defined by the dynamic equations \eqref{eq:er_dyn_st_sp} is expected to follow a trajectory that is as close to the real-world trajectory as possible for the same set of control inputs with constant longitudinal velocity $V_x$. For the real world performance, according to the feedback controller implementation in \cite{rajamani2011vehicle} the steady state values of $e_1$ and $e_2$ may be non-zero but should converge close to zero.

\subsection{Model Identification}
The problem formulation becomes an optimization problem of minimizing the gap between the simulated trajectory and the real robot trajectory as follows:
\begin{equation}
\begin{aligned}\label{eq:gain_tuning}
    \min_{\hat{P}} \mathcal{\hat{L}}(r,\hat{r})
\end{aligned}
\end{equation}

where $\mathcal{\hat{L}}$ representing the gap between the real and simulated trajectories $\hat{r}$ and $r$ respectively.

\subsection{Verifying the Control Law in Real World}

 The open-loop matrix $A$ may have eigenvalues at the origin and be unstable. Using the state feedback law, the eigenvalues of the closed-loop matrix $(A-B\boldsymbol{K})$ can be placed at desired locations. The closed-loop system, as defined in Definition \ref{def:closed_loop}, using this state feedback controller is therefore:
\begin{equation}
\begin{aligned}\label{eq:stabilized_AB}
    \dot{x} = (A-B_1\boldsymbol{K})x + B_2 \dot{\Psi}_{des}
\end{aligned}
\end{equation}
To compute the feedback matrix $\boldsymbol{K}$, the pole placement algorithm \cite{kautsky1985robust} is used.
\begin{equation}
\begin{aligned}\label{eq:pole_placement}
    \boldsymbol{K} = place(A, B_1, P)
\end{aligned}
\end{equation}

Where the vector $P$, which is user defined, defines the desired pole locations such that the eigenvalues of the matrix $(A-B_1K)$ are negative, non-zero real numbers, satisfying the \textit{full state feedback} system stability condition. From Eq.~\eqref{eq:er_dyn_st_sp} it is observable that the matrices $A$ and $B_1$ are defined by the front-steered vehicle dynamic parameters $m$, $l_f$, $l_r$, $C_{af}$ and $C_{ar}$.
% \btxt{Maybe add such that for the pole placement under the optimization formulation?}
We hypothesize that correctly identifying the set of system dynamic parameters in Eq.~\eqref{eq:er_dyn_st_sp} would yield the gain vector $\boldsymbol{K}$ that would make the steady state values of $e_1$ and $e_2$ converge close to zero in real world experiments.

%% file: 4-proposed_method.tex
\section{Method}\label{Proposed Method}

\subsection{Vehicle Data Collection}\label{sec:Vehicle Initialization and Trajectory Design}

\begin{figure}
\centering
\subfloat{\includegraphics[width=0.48\textwidth]
{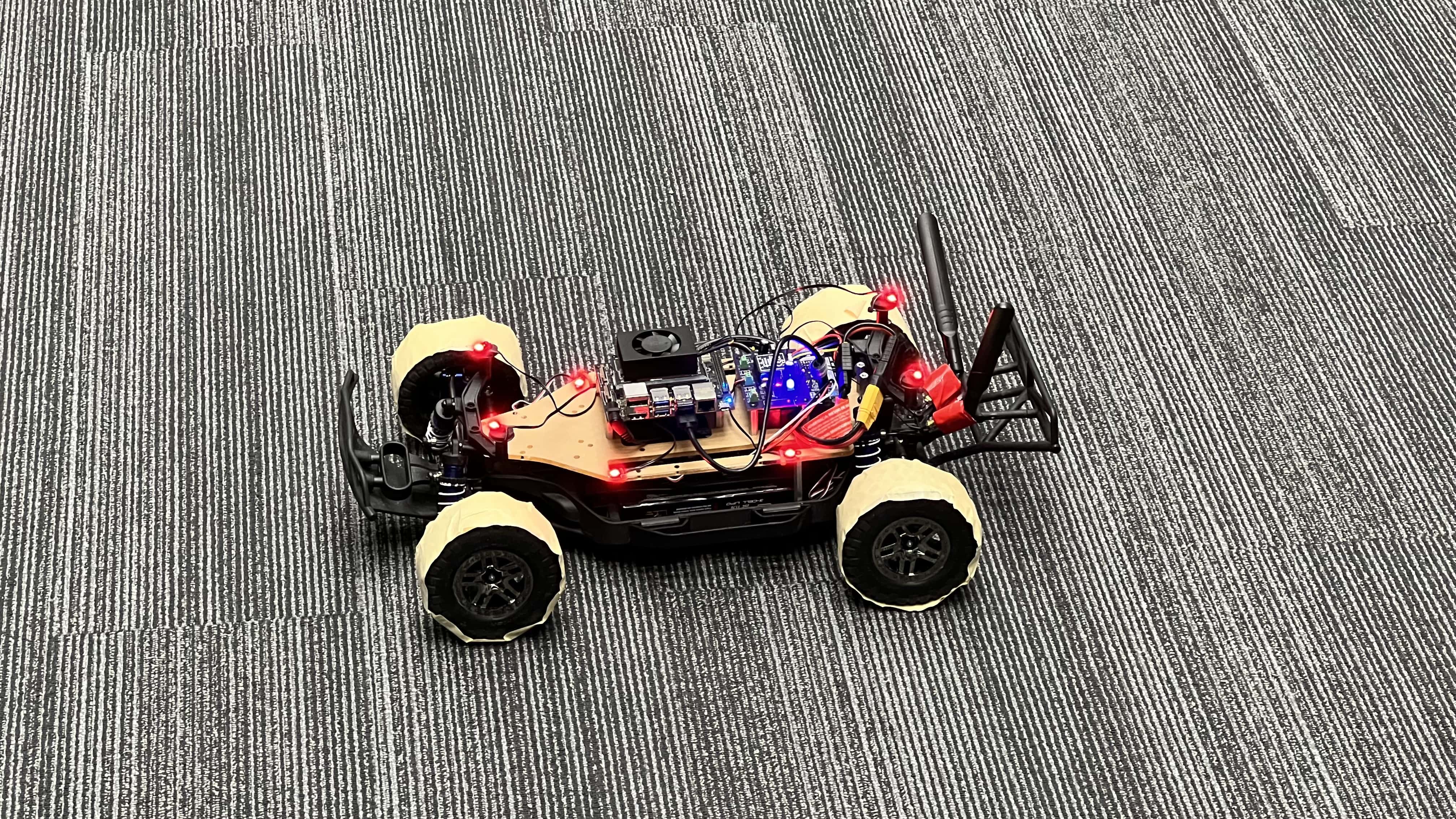}}
\caption{F1TENTH with a lower tire friction configuration. The rubber tires are wrapped using a tape that has less grip on carpet surface.} 

\label{fig:taped_vehicle}
\end{figure}

The vehicle was driven indoors with Phasespace X2E LED motion capture markers attached on top of it for 6 DoF state estimation. The vehicle was autonomously operated through its ROS interface \cite{quigley2009ros} for 6-10 seconds. To test how general our method is, we also wrapped the vehicle tires with tape for a second set of experiments, which is expected to reduce the friction between the vehicle tires and the ground (Fig. \ref{fig:taped_vehicle}). For simplicity, a total of 16 circular trajectories were collected for training equally split between left and right turns at maximum steering angle at $1 \ m/s$ velocity for both friction conditions. To preserve compatibility with the state-space definition provided in Eq. \eqref{eq:er_gnrl_st_sp}, we do not apply any steering velocity or longitudinal acceleration commands which is represented as $u_{1} = 0$ and $u_{2} = 0$. Instead, we are directly modifying the states $x_{3}$ and $x_{4}$, which are the steering angle and longitudinal velocity, respectively.

\subsection{Loss Function}
The loss function needs to minimize the gap between the simulated and the real robot trajectory:
\begin{equation}
\begin{aligned}\label{eq:Ldtw}
    \mathcal{L}_{dtw} = dtw(r,\hat{r})
\end{aligned}
\end{equation}

where $dtw$ refers to the differentiable implementation of Dynamic Time Warping \cite{muller2007dynamic, maghoumi2020dissertation, cuturi2017soft} algorithm. In practice, Dynamic Time Warping does not penalize the scale difference between trajectories enough, therefore it needs to be combined with another loss term to prevent trajectories of similar shapes but large scale differences. For that purpose, we use the chamfer distance as the second loss term.

\begin{equation}
\begin{aligned}\label{eq:Ldtw}
    \mathcal{L}_{cd} = cd(r,\hat{r})
\end{aligned}
\end{equation}
A weighted linear combination of $\mathcal{L}_{dtw}$ and $\mathcal{L}_{cd}$ is proposed as the actual loss function:

\begin{equation}
\begin{aligned}\label{eq:Ltot}
    \mathcal{L} = \mathcal{L}_{dtw} + \lambda\mathcal{L}_{cd}
\end{aligned}
\end{equation}

where the weight is empirically determined as $\lambda = 100$.

%% file: 5-simulation_results.tex
\section{Simulation Results}\label{Simulation Results}

In this section, we first compare our gradient-based method to a gradient-free baseline. The front-steered vehicle was simulated through our differentiable physics engine with  timestep size of $0.002 \ s$. Our simulator implements the nonlinear dynamics given in Eq. \eqref{eq:adp_vdot} and adopted by the official F1TENTH simulator \cite{o2020f1tenth}. $h_{cg}$ is fixed as 0.074 meters, $I_{z}$ is fixed as $0.04712 \ kg.m^2$ m is measured at $3.1 \ kg$. The coefficient of friction between the vehicle wheels and the ground was set to $\mu = 1.0489$ and the steering angles were clipped at each timestep, to remain in the interval: $[-0.34, 0.34]\ rads$ in accordance with the physical model properties. The velocity was kept constant at $1 \ m/s$. In our experiments, we are looking to identify the following model parameters: $l_{f}$, $l_{r}$, $C_{S,f}$, $C_{S,r}$. For training, batch size is set to 4 for Adam optimizer \cite{kingma2014adam}. Loss and parameter figures are plotted as the average of $5$ separate training rounds with uniformly sampled random initial parameters. For the baseline, gradient-free optimizer CMA-ES \cite{hansen2016cma} is used for optimizations from the Optuna library \cite{akiba2019optuna}.

\begin{figure}
\centering
\subfloat[]{\includegraphics[width = 0.5\linewidth]{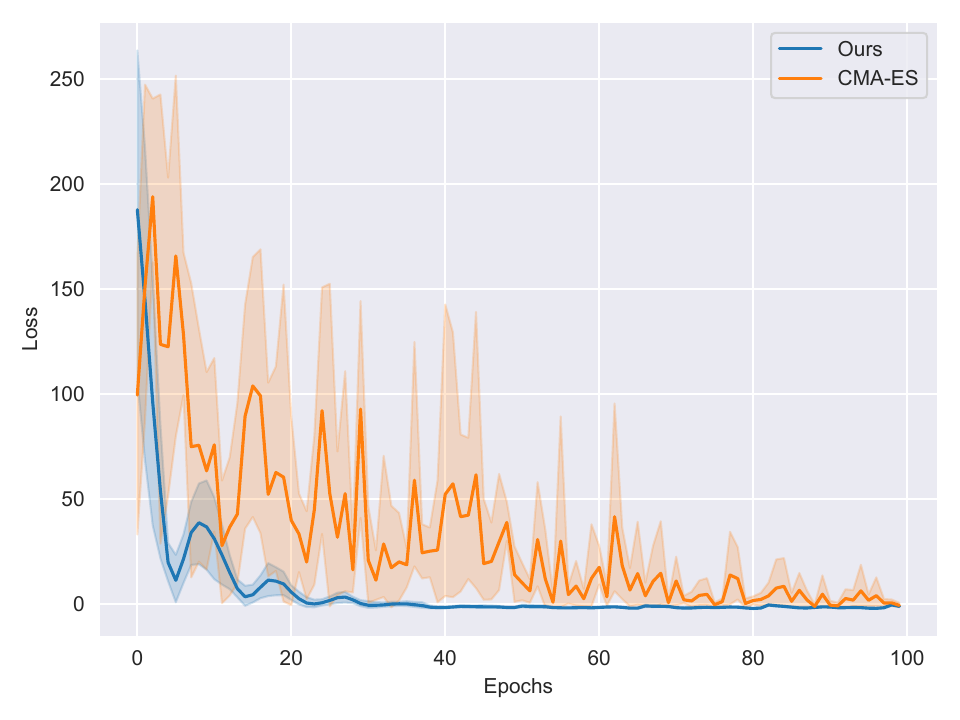}\label{fig:untapedgbgf}}
\subfloat[]{\includegraphics[width = 0.5\linewidth]{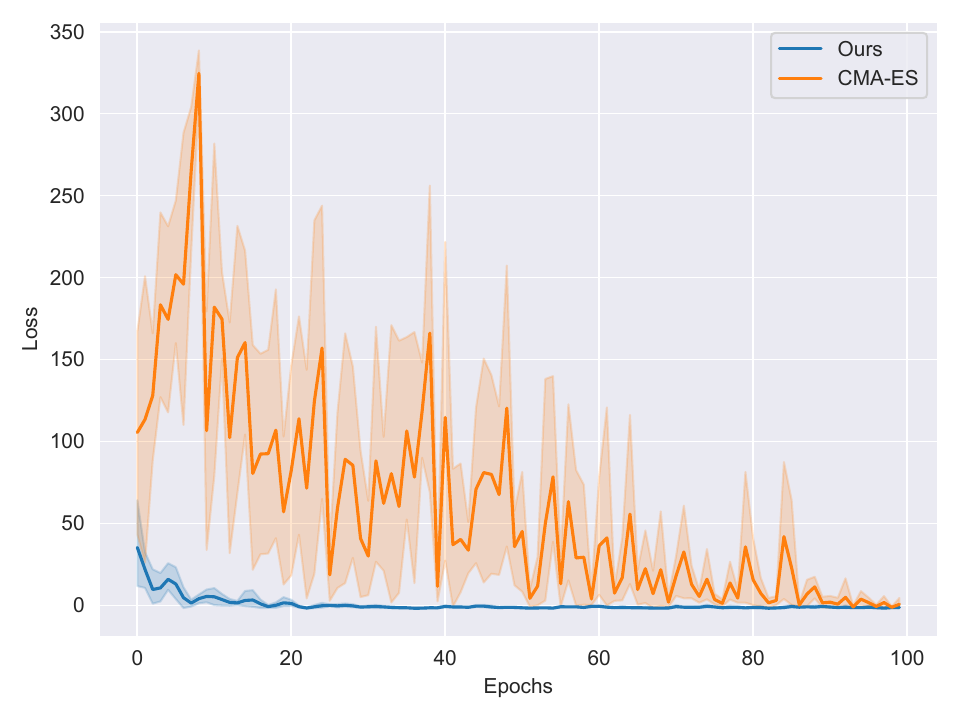}\label{fig:tapedgbgf}}

\caption{The loss trend for model identification. Both methods use the analytical methods but only our method makes use of the analytical gradients provided by our differentiable physics engine. Analytical gradients provide higher sample efficiency and the training converges in far fewer iterations, (a) shows the model identification of the F1TENTH vehicle with standard tire configuration (b) shows the model identification for the lower friction configuration.}
\label{fig:loss_results}
\end{figure}

We present the loss curves for gradient-based and gradient-free gain optimizations in Fig. \ref{fig:loss_results} for 100 epochs. Compared to the gradient-free baseline, our gradient-based method provides much higher sample-efficiency and a more regular loss curve enabled by the analytical gradients for standard and lower friction tire configurations both. The effect of the analytical gradients is particularly evident in the first few epochs where gradient-based methods have to initially explore the search space. In Table \ref{tab:modelParams} we present the model parameters estimated by the gradient-based and gradient-free optimization methods averaged over 5 initializations sampled uniformly at random. The first two rows demonstrate the values provided by the official implementation of the F1TENTH vehicle and our method's results, respectively. The results suggest that our method is able to closely match more sophisticated system identification methods through a simpler procedure. Per Eq. \eqref{eq:adp_vdot}, for a simulation with fixed friction coefficient, our hypothesis was to observe a significantly lower cornering stiffness coefficient for the lower friction configuration. In the third row of Table \ref{tab:modelParams} our method's results are presented. They are a bit unexpected as the cornering stiffness values are higher. We hypothesize that experiments using acceleration and steering velocity commands would get closer to the friction limits of the vehicle would make this change of parameters more observable.

\begin{table}[]
\small
\begin{center}
\resizebox{\linewidth}{!}{
\begin{tabular}{|c|c|c|c|c|}
\hline
 \textbf{Estimation Method} & $l_f$ & $l_r$ & $C_{S,f}$ &  $C_{S,r}$\\ \hline
 True &  $0.159$ & $0.171$  & $4.728$ & $5.546$ \\ \hline
Ours & $0.142$ & $0.171$ & $5.909$ &  $4.767$\\ \hline
 Ours(low friction)&  $0.161$ & $0.143$& $5.113$ &  $4.988$\\ \hline
 CMA-ES&  $0.127$ & $0.194$& $7.442$ &  $6.147$\\ \hline
 CMA-ES(low friction)&  $0.155$ & $0.143$& $6.823$ &  $8.960$\\ \hline
\end{tabular}}
\caption{ESTIMATED MODEL PARAMETERS}
\label{tab:modelParams}
\end{center}
\end{table}

%% file: 6-generalization_experiments.tex
\section{Real World Experiment Results}\label{Generalization Experiment Results}
\begin{figure}[]
\centering

\subfloat[]{\includegraphics[width = 0.5\linewidth]{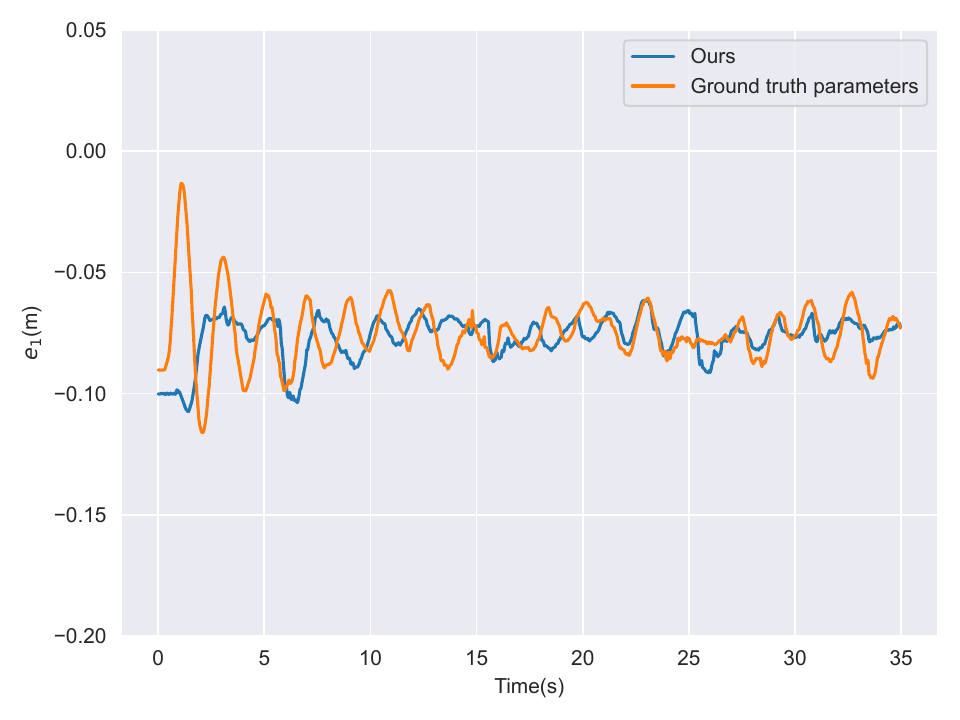}\label{fig:untaped_errors}}
\subfloat[]{\includegraphics[width = 0.5\linewidth]{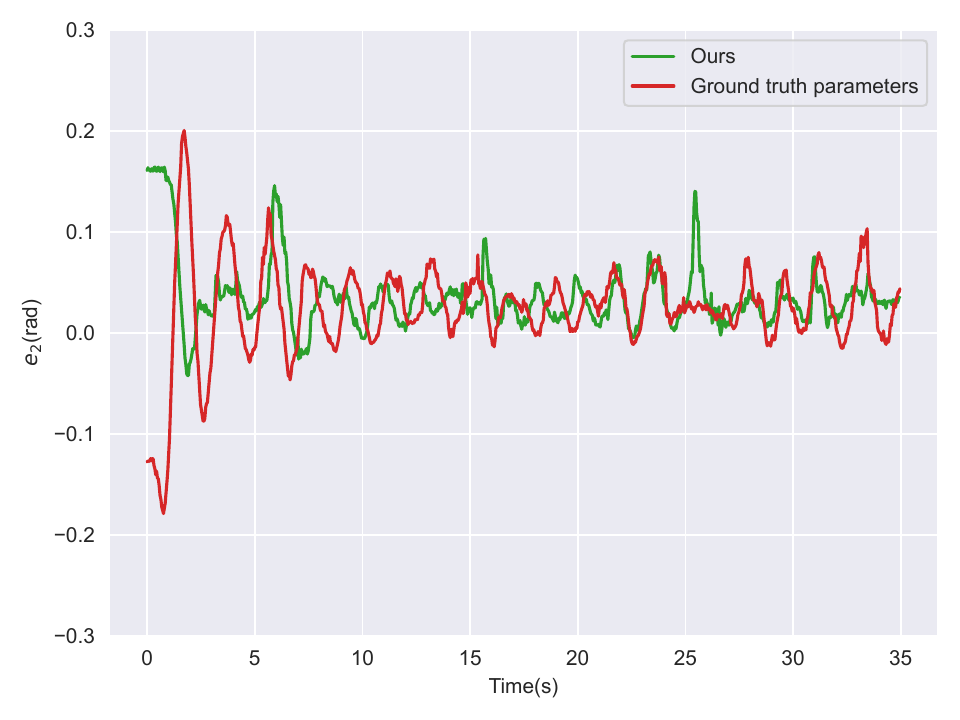}\label{fig:untaped_errors2}}

\subfloat[]{\includegraphics[width = 0.5\linewidth]{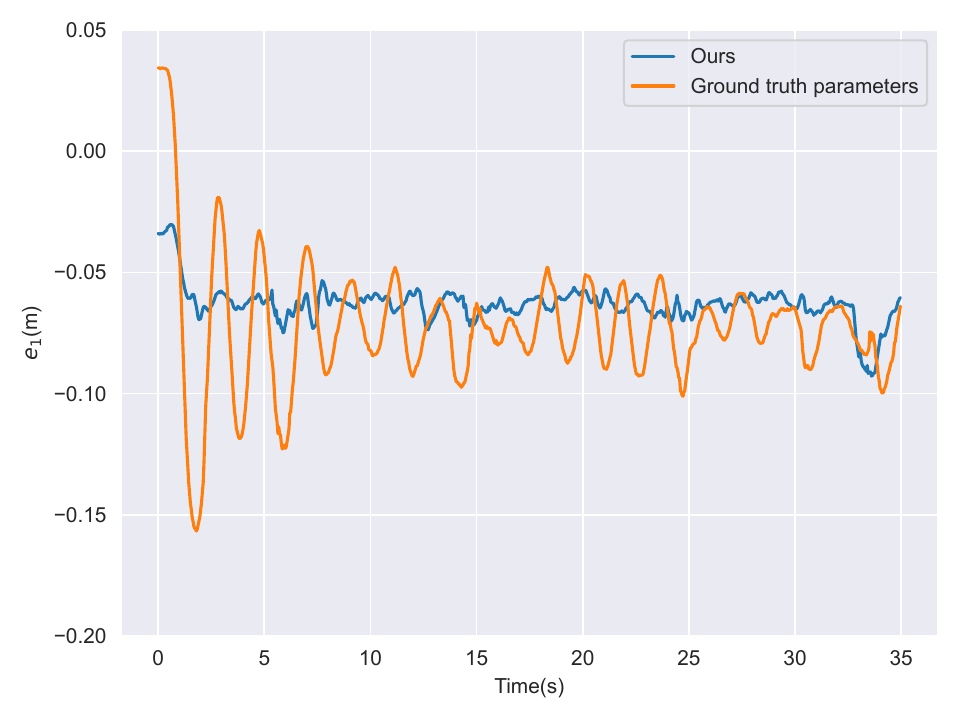}\label{fig:taped_errors}}
\subfloat[]{\includegraphics[width = 0.5\linewidth]{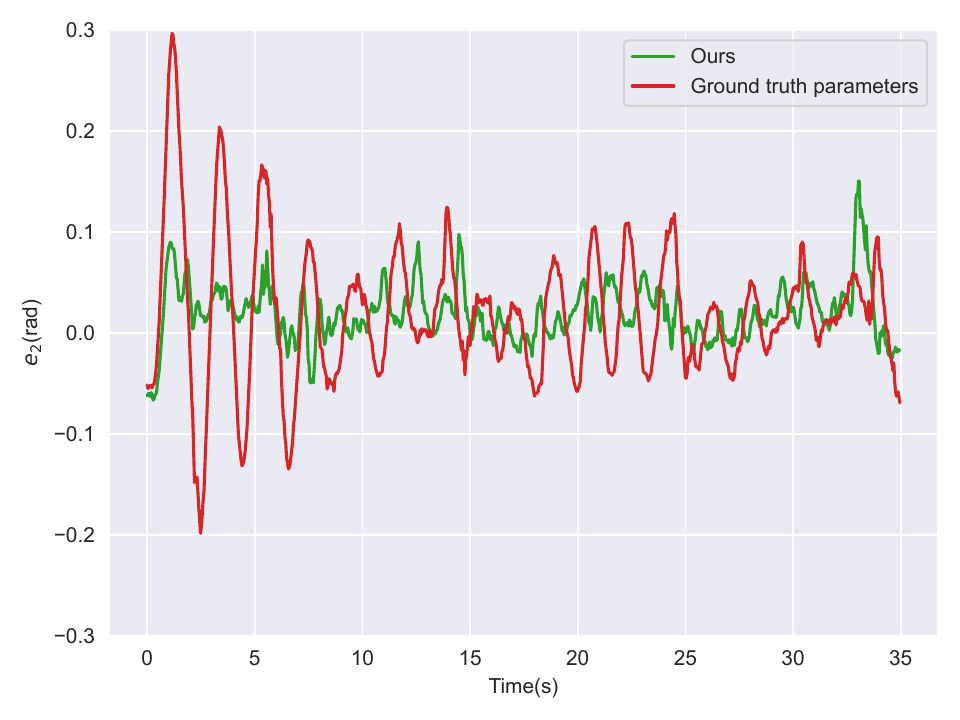}\label{fig:taped_errors2}}

\caption{Error profile of $e_1$ in meters and $e_2$ in radians for (a) and (b) F1TENTH vehicle with standard tire configuration (c) and (d) with the lower friction configuration. In both cases, error profiles confirm that the feedback controller using parameters identified through our method has comparable performance w.r.t the ground truth parameters identified in the official implementation \cite{o2020f1tenth}.}
\label{fig:ep_results}
\end{figure}

For the real world experiments, the feedback controller presented in Eq. \eqref{eq:stabilized_AB} is used. The poles were set as $P=[-2+2j,-2-2j,-150+15j,-150-15j]$ since left hand side pole placement is necessary for the stability of the controller. The error profiles presented in Fig. \ref{fig:ep_results} for $e_1$ and $e_2$ demonstrate that the identified models are transferable to real systems and optimized controllers show similar performance. As discussed in Section \ref{Simulation Results}, two set of experiments were conducted. In \ref{fig:untaped_errors} and \ref{fig:untaped_errors2}, for the standard tire configuration case, the error profiles demonstrate comparable performance to the ground truth parameters identified in the official implementation with the identified system parameters in our experiments. For the lower friction case, in \ref{fig:taped_errors} and \ref{fig:taped_errors2}, the identified model for the lower friction case performs well, however the identified standard parameters do not seem to suffer significantly and the controller performance is still robust. We attribute the robustness of the default controller performance under lower friction setting to the relatively conservative longitudinal velocity and feedback mechanism's ability to compensate small model identification errors, evident in Sec \ref{Simulation Results}, where identified model parameters are relatively similar. For further details on our experimental results, please refer to the submitted video attachment.
% width = 0.8\linewidth
% width=9cm,height=5cm

%% file: 7-Conclusion_and_future_directions.tex
\section{Conclusions and Future Directions }\label{Conclusions and Future Directions}

Front-steered vehicles constitute the majority of driving equipment. In this paper we presented a method for system-identification and control of front-steered vehicles and demonstrated our approach on the F1TENTH vehicle, which abides by the Ackerman geometry constraints, using a differentiable physics engine combined with gradient-based optimization methods. With the help of the analytical gradients provided by the differentiable physics engine, our method converges in far fewer iterations compared to the gradient-free baseline. Our proposed method identifies the unknown parameters of the system as well as executes a stable feedback controller used to achieve lane keeping. We provided experimental results using an F1TENTH vehicle exhibiting comparable lane keeping behavior for system parameters learned using our gradient-based method with lane keeping behavior of the \textit{actual} system parameters of the F1TENTH.

For the future, we will work on implementing an online system identification and controller methodology for front-steered Ackermann vehicles. %We will also explore more articulated control architectures.

% It is particularly critical that such a framework exists because front-steered vehicles make majority of the on-road and off-road vehicles we interact with everyday. 
% Testing our framework on real robots remains as our future work. We also intend to incorporate more sophisticated controllers such as \textit{robust} and \textit{adaptive} controllers.